\documentclass[letterpaper, 10 pt, conference]{ieeeconf}  

\IEEEoverridecommandlockouts                              

\usepackage{amsmath,amsfonts}
\usepackage{graphicx}
\usepackage{gensymb}
\usepackage{textcomp}
\usepackage{cite}

\title{\LARGE \bf
Wrist-Squeezing Force Feedback Improves Accuracy and Speed in Robotic Surgery Training
}

\author{Sergio Machaca$^{1}$, Eric Cao$^{2}$, Amy Chi$^{3}$, Gina Adrales MD MPH$^{4}$, \\Katherine J Kuchenbecker PhD$^{5}$, Jeremy D Brown PhD$^{6}$
\thanks{$^{1}$Sergio Machaca is with Department of Mechanical Engineering, Johns Hopkins University, Baltimore, MD 21218, USA
        {\tt\small smachac2@jh.edu}}%
\thanks{$^{2}$Eric Cao was with the Department of Biomedical Engineering, Johns Hopkins University, Baltimore, MD 21218, USA}%
\thanks{$^{3}$Amy Chi was with the Department of Mechanical Engineering, Johns Hopkins University, Baltimore, MD 21218, USA}%
\thanks{$^{4}$Gina Adrales is with the Johns Hopkins University School of Medicine, Baltimore, MD 21205, USA}%
\thanks{$^{5}$Katherine J Kuchenbecker is with the Max Planck Institute for Intelligent Systems, Stuttgart, Germany}%
\thanks{$^{6}$Jeremy D Brown is with the Department of Mechanical Engineering, Johns Hopkins University, Baltimore, MD 21218, USA}%
}

\begin{document}

\maketitle
\thispagestyle{empty}
\pagestyle{empty}

\begin{abstract}

Current robotic minimally invasive surgery (RMIS) platforms provide surgeons with no haptic feedback of the robot's physical interactions. This limitation forces surgeons to rely heavily on visual feedback and can make it challenging for surgical trainees to manipulate tissue gently. Prior research has demonstrated that haptic feedback can increase task accuracy in RMIS training. However, it remains unclear whether these improvements represent a fundamental improvement in skill, or if they simply stem from re-prioritizing accuracy over task completion time. In this study, we provide haptic feedback of the force applied by the surgical instruments using custom wrist-squeezing devices. We hypothesize that individuals receiving haptic feedback will increase accuracy (produce less force) while increasing their task completion time, compared to a control group receiving no haptic feedback. To test this hypothesis, N=21 novice participants were asked to repeatedly complete a ring rollercoaster surgical training task as quickly as possible. Results show that participants receiving haptic feedback apply significantly less force (0.67 N) than the control group, and they complete the task no faster or slower than the control group after twelve repetitions. Furthermore, participants in the feedback group decreased their task completion times significantly faster (7.68\%) than participants in the control group (5.26\%). This form of haptic feedback, therefore, has the potential to help trainees improve their technical accuracy without compromising speed.

\end{abstract}

\section{INTRODUCTION}

Robotic minimally invasive surgery (RMIS) is a frequently used method of treatment for many routine and non-routine surgical procedures \cite{McGuinness2018RoboticsUrology,Sheetz2020TrendsProcedures}. By providing enhanced visualization and dexterity, RMIS platforms aim to overcome the limitations of laparoscopic surgery, which include limited range of motion, poor surgical tool ergonomics, and a two-dimensional video feed \cite{Ballantyne2002TheSurgery}. The most commonly used RMIS platform, the Intuitive da Vinci surgical system, has yielded promising surgical outcomes 
in several procedures such as prostatectomy, nephrectomy, transoral surgery, and hernia repair \cite{McGuinness2018RoboticsUrology,Ballantyne2007TeleroboticEfficacy,Weinstein2012TransoralMargins,Patel2007RoboticCases,Soliman2020Robot-assistedRepair,Donkor2017CurrentRepair}. The same platform has also sometimes been criticized for incurring high cost and slow procedure times without delivering substantially better clinical outcomes \cite{Ahmad2017RoboticEvidence,Kang2011ConventionalAdvantages}. \looseness=-1

Despite the advantages of RMIS, teleoperation of the surgical instruments removes the surgeon's ability to feel the physical interactions between the instruments and the surgical environment \cite{Okamura2009}. The lack of haptic information causes surgeons to rely solely on visual feedback to estimate applied instrument forces and tissue properties. This limitation can present significant challenges to surgeons trying to complete tasks such as palpating tissue, tying sutures, or localizing visually occluded target anatomy.

In an attempt to solve the haptic feedback limitation, significant research has investigated the utility of adding haptic feedback to RMIS platforms. Many studies have demonstrated the benefits of various kinds of haptic feedback in RMIS. For example, Koehn and Kuchenbecker showed that surgeons and non-surgeons prefer haptic feedback of instrument vibrations when performing surgical tasks \cite{Koehn2015a}, as opposed to no haptic feedback (visual cues only). Similarly, King et al.\ showed that tactile feedback significantly reduced the grip force applied by both novice and expert surgeons when performing a peg transfer task \cite{King2009TactileSurgery}. Using a bilateral teleoperator in a vessel dissection task, Wagner et al.\ demonstrated that force feedback significantly reduced the magnitude of the instrument tip forces applied by non-surgeons, medical students, surgical residents, and attending surgeons alike \cite{Wagner2007TheDissection}. On the contrary, there have been several studies that have suggested little to no perceived benefit of haptic feedback (see \cite{VanDerMeijden2009TheReview} for a review). As suggested by Hagen et al., visual cues can act as a substitute for haptic feedback \cite{Hagen2008}, reducing its overall utility. It is likely that these conflicting viewpoints on the utility of haptic feedback for RMIS have limited the translation of haptic feedback to clinical use. \looseness=-1

Given the demonstrated utility of haptic feedback in laparoscopic training \cite{Zhou2012EffectAcquisition,Wottawa2013TheTasks}, dental training \cite{Kuchenbecker2017EvaluationDetection,Al-Saud2017FeedbackSimulator}, and teleoperation more broadly \cite{Khurshid2017EffectsTask,Pacchierotti2015CutaneousSystems}, several researchers have explored the use of haptic feedback for RMIS training \cite{Brown2017,VanDerMeijden2009TheReview,Coles2011TheArt}. Brown et al.\ for example demonstrated that wrist-squeezing haptic feedback of interaction force leads to improved accuracy (reduced contact forces) in an inanimate ring rollercoaster training task for novice da Vinci users \cite{Brown2017}; notably, the performance improvement was sustained even after the haptic feedback was removed. Likewise, Abiri et al.\ demonstrated a significant reduction in grip force for both novice and expert surgeons using a pneumatic force-feedback system, noting that the presence of feedback had an even more significant impact on novices \cite{Abiri2019Multi-ModalSurgery}. These benefits merit further exploration to determine how accuracy is impacted when training with haptic feedback, compared to conventional learning without feedback, for the same task.

It is also worth considering, from a skill acquisition standpoint, how these increases in performance affect task completion time. The tradeoff between speed and accuracy in motor tasks is well known \cite{Meyer1982ModelsMovements} and has been investigated in the context of surgery more broadly. A few studies have demonstrated that a tradeoff exists between accuracy and task completion time for robotic surgery, wherein increases in accuracy come at the expense of longer task completion times \cite{Chien2010AccuracySurgery}. There have also been studies that have demonstrated that haptic feedback improves accuracy and task completion time \cite{Bark2012SurgicalTasks, Abiri2019ArtificialFeedback}. However, it is not always implied that these improvements in accuracy and task completion time are superior to conventional manipulation (without haptic feedback). When someone increases either their speed or their accuracy, the other metric often decreases, yielding a similar overall level of performance; deeper learning of the skills required, however, can improve both simultaneously. Regarding haptic feedback in RMIS, Brown et al.\ found that their wrist-squeezing haptic feedback increased task accuracy but had a mixed impact on task completion time, with some participants performing the task faster and others performing the task slower \cite{Brown2017}. Also, their wrist-squeezing devices squeezed both wrists at the same time when either of the instruments were in contact with the task materials, providing no information about handedness. Using a bimodal vibrotactile system to convey applied forces, Abiri et al.\ found that haptic feedback led to significantly faster task completion times in a vessel localization task \cite{Abiri2019ArtificialFeedback}; force accuracy was not reported. Based on these different findings, we are interested in understanding more generally how the speed-accuracy tradeoff differs for trainees receiving bimanual haptic feedback (with handedness information) and those that are not. 

\begin{figure}[t]
    \centering
    \includegraphics[width=0.8\columnwidth]{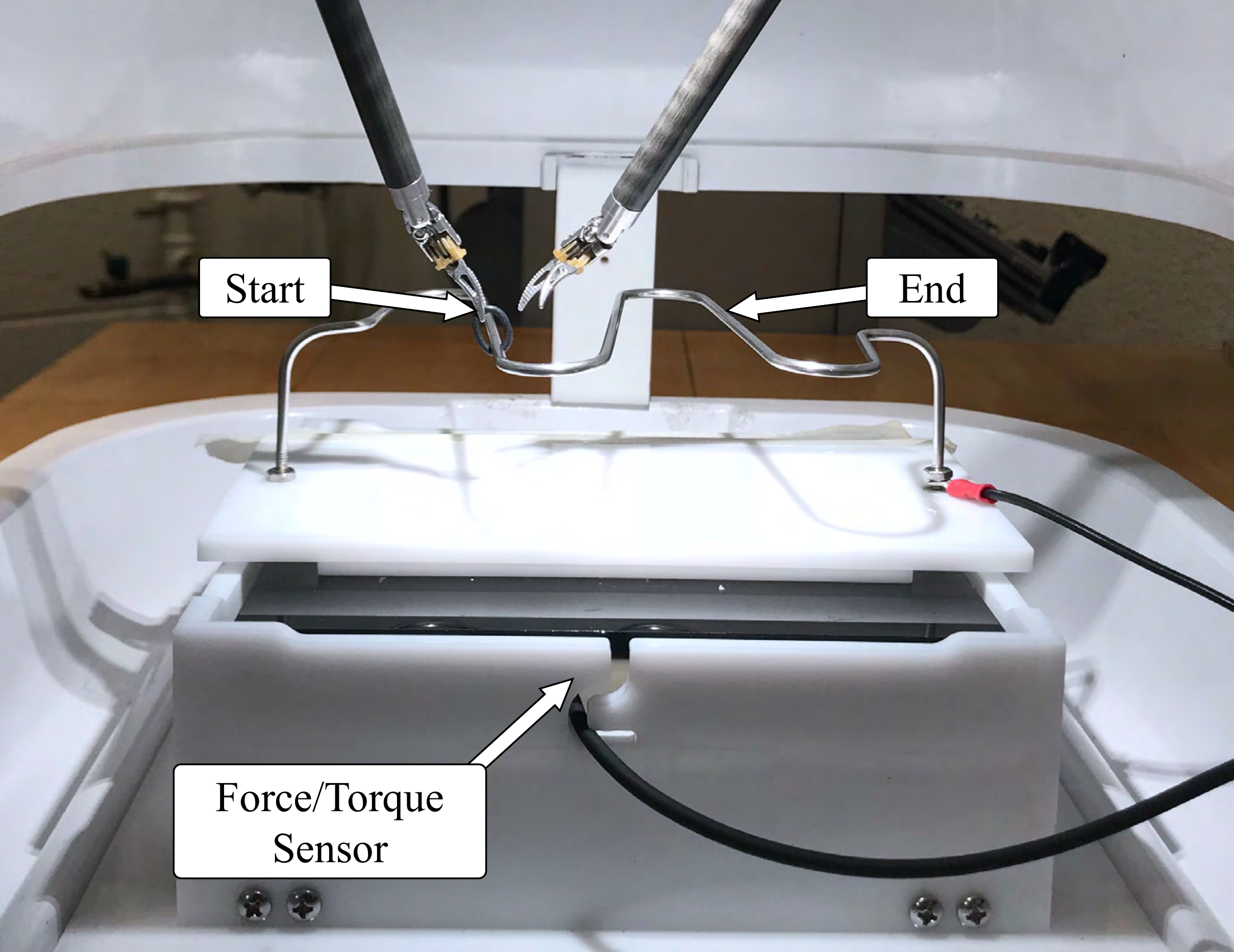}
    \caption{Ring rollercoaster training task magnetically attached to the instrumented housing containing a six-axis force/torque sensor. The start and end positions used during the experiment are labeled.}
    \label{fig:TASK_PLATFORM}
    \vspace{-10pt}
\end{figure} 

While prior research has demonstrated that haptic feedback can increase task accuracy in RMIS training, it remains unclear whether these improvements represent a fundamental improvement in skill, or if they simply stem from re-prioritizing accuracy over task completion time. Furthermore, no prior studies have analyzed the rate at which the task completion time changes over repeated trials of an RMIS training task when receiving haptic feedback, nor whether that rate is different from conventional learning. In this study, we investigate the impact of bimanual haptic feedback on the speed-accuracy tradeoff in RMIS training compared to conventional learning. Using a between-subjects design, we compare performance in terms of accuracy (interaction force magnitude) and task completion time in the inanimate ring rollercoaster training task between novice da Vinci participants receiving wrist-squeezing haptic feedback of interaction force magnitude and a control group receiving no haptic feedback. Because they can physically perceive the consequences of instrument contact, we hypothesize that participants who receive haptic feedback will prioritize accuracy over speed, reducing their interaction forces while increasing their task completion time compared to the control group.

\section{Methods}

\subsection{Experimental Setup}

Our setup consists of an Intuitive Surgical da Vinci robot, an instrumented training platform, and a bimanual haptic feedback system. This setup is based on the experimental setup first presented by Brown et al. \cite{Brown2017}.

\subsubsection{da Vinci Robot}

The da Vinci S HD surgical system consists of three main subsystems: patient-side cart, vision cart, and surgeon's console. We instrumented the da Vinci with two Maryland bipolar forceps. The motion scale factor was set to normal (2:1 scale factor) and the 30-degree scope was set to the ``down'' configuration. Throughout the experiment, participants were not allowed to move the camera, zoom the camera in or out, or use the clutch pedal to adjust the workspace of the da Vinci manipulators.

\subsubsection{Instrumented Training Platform}

A ring rollercoaster training task (Intuitive Surgical da Vinci Skills Drill Practicum) is magnetically attached to a custom acrylic platform housing an ATI Mini40 six-axis force/torque sensor (see Fig. \ref{fig:TASK_PLATFORM}). The instrumented training platform is centered within a white da Vinci skills dome, which simulates the human abdomen and features various trocar ports. A custom handedness circuit detects contact between each surgical instrument and the task materials. The circuit consists of a separate voltage divider for each instrument; it utilizes the electrical conductivity of the metal rollercoaster track, a conductive o-ring (Marco Rubber S1104-010), and the two bipolar Maryland forceps to distinguish between left and right instrument contacts.

\subsubsection{Bimanual Haptic Feedback System}

\begin{figure}
    \centering
    \includegraphics[width=0.8\columnwidth]{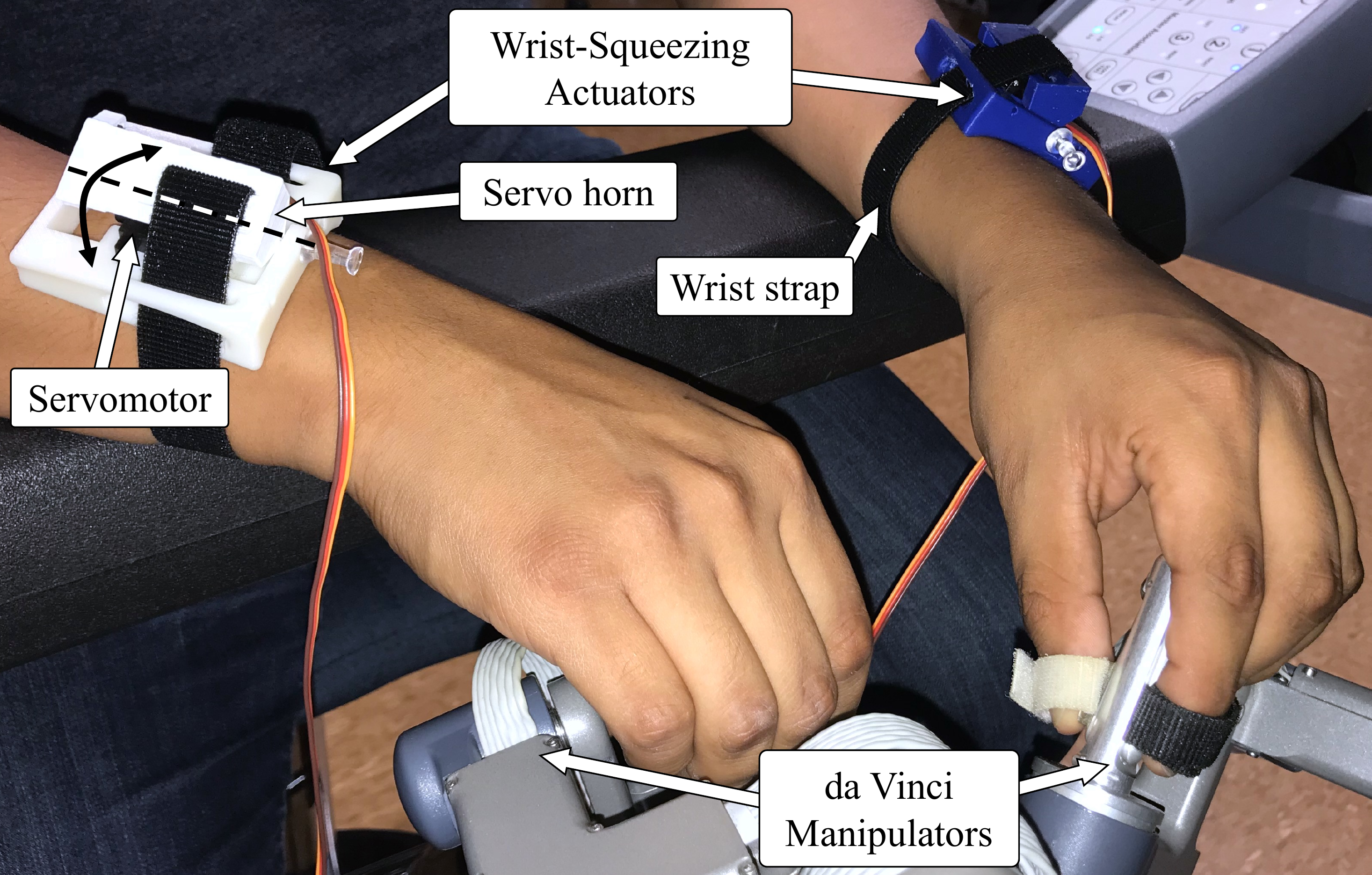}
    \caption{Participant wearing the bimanual wrist-squeezing tactile actuators. Each actuator is controlled independently. The hook-and-loop strap is fixed to the 3D-printed frame on one end and attached to the servomotor horn on the other end. As the force applied by an instrument increases, the corresponding servo angle increases, and the hook-and-loop strap tightens around the participant’s wrist creating a squeezing sensation.}
    \label{fig:WRIST_SQUEEZERS}
    \vspace{-10pt}
\end{figure}

As seen in Fig.~\ref{fig:WRIST_SQUEEZERS}, our bimanual haptic feedback system consists of two wrist-squeezing tactile actuators based on work by Stanley and Kuchenbecker \cite{Stanley2012}. Each actuator consists of a Futaba s3114 servomotor (maximum output torque: 0.17\,Nm) housed in a 3D-printed frame. Each frame is fastened on one of the participant's wrists with a hook-and-loop strap, which is connected to the frame on one end and secured to the servo horn on the other end. The servomotors produce a squeezing sensation on participants' wrists by shortening the strap as it rotates through its range of motion. A servo driver (Phidgets 1061\_0) drives the servomotor angle in proportion to the interaction forces measured by the force/torque sensor according to:
\begin{equation}
    \theta = 
    \begin{cases}
        \theta_{\min} & \text{if} \hspace{5pt} F \leq F_{\text{th}} \\
        k (F - F_{\text{th}}) (\theta_{\max} - \theta_{\min}) & \text{if} \hspace{5pt} F_{\text{th}} < F \leq 2\,\text{N} \\
        \theta_{\max} & \text{if} \hspace{5pt} F > 2\,\text{N}
    \end{cases}
    \label{SERVO_ANGLE_EQN}
\end{equation}
\noindent where $\theta$ is the desired angle sent to the servo, $F$ is the Euclidean magnitude of the 3D force vector measured by the force/torque sensor, $F_{\text{th}}$ = 0.2\,N is a threshold magnitude used to reduce noise at low forces, and $k$ = 0.56\,N$^{-1}$ is a linear scaling factor that maps force to position. Larger angles squeeze the wrist more firmly. $\theta_{\max}$ = 50$\degree$ and $\theta_{\min}$ = 0$\degree$ are the maximum and minimum angles to which the servos are commanded. The servo remains at its maximum angle when the force signal exceeds 2\,N; this limit and the other parameters of this haptic feedback mapping were determined empirically through pilot trials. The torque measurement from the force/torque sensor is not utilized in the force magnitude calculation.

Utilizing the handedness circuit, each wrist-squeezing actuator responds if the corresponding da Vinci instrument is making contact with the task materials. For instance, if the participant touches the ring rollercoaster track with the left instrument, then the actuator on the participant's left arm tightens the wrist strap in proportion to the interaction force. If both instruments contact the task track, or if neither instrument contacts the track when a force is being applied, tactile feedback is delivered equally to both wrists. 

\subsubsection{Data Acquisition and Control}

\begin{figure}
    \centering
    \includegraphics[width=0.8\columnwidth]{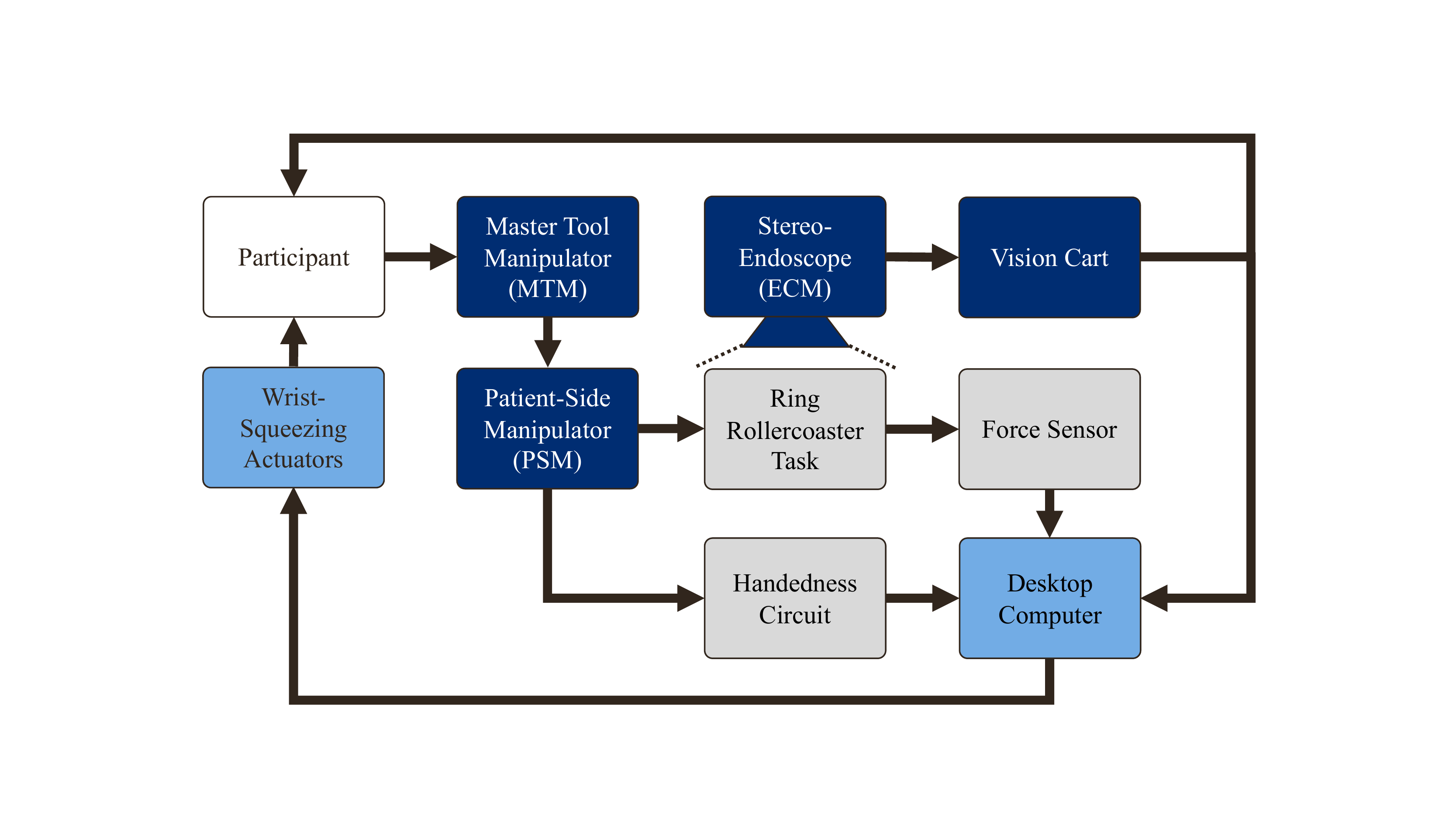}
    \caption{Experimental setup flow diagram. Components of the da Vinci are shown in dark blue, instrumented training platform in gray, and bimanual haptic feedback system in light blue.}
    \label{fig:SETUP_FLOW}
    \vspace{-5pt}
\end{figure}

We used a custom data acquisition (DAQ) board containing chipsets for filtering and analog-to-digital conversion. The DAQ board receives six channels of raw voltage data from the ATI Mini40, which are used to calculate the force along and torque about the x, y, and z axes. An onboard Teensy 3.2 microcontroller buffers the signal and communicates with an Intel NUC computer via USB. On the computer, a custom Python script receives serial packets from the Teensy and converts raw voltage data into forces and torques, recording at a frequency of 50 Hz. In addition to the sensor readings from the force/torque sensor and the handedness circuit, the endoscopic video is recorded a rate of approximately 30 fps. Using the measured interaction forces, the Python script controls the servomotor angle according to \eqref{SERVO_ANGLE_EQN}. The flow of communication between hardware systems is shown in Figure~\ref{fig:SETUP_FLOW}.

Participants start and stop the recording of each trial using a foot pedal on the surgeon's console. The Python script receives the pedal input and tares the force sensor before each trial begins.

\subsection{Experiment Protocol}

\subsubsection{Participants}

We recruited N=21 novice participants (10 male, 11 female, mean age 19.2$\pm$1.2 years) from the adult population of the Johns Hopkins University. All participants were between 18 and 22 years old. From the initial pool of 21 participants, one was removed from the dataset since the ring rollercoaster track was not securely attached to the platform during their trials, which created erroneous data. Of the remaining N=20 participants, 18 had no prior experience using the da Vinci surgical system, and two had used it once in a demonstration. All participants were consented according to a protocol approved by the Johns Hopkins Homewood Institutional Review Board (\#HIRB00005942) and were compensated at a rate of \$15 per hour. The experiment lasted approximately one hour. 

\subsubsection{Experiment Design}

\begin{figure}
    \centering
    \includegraphics[width=0.8\columnwidth]{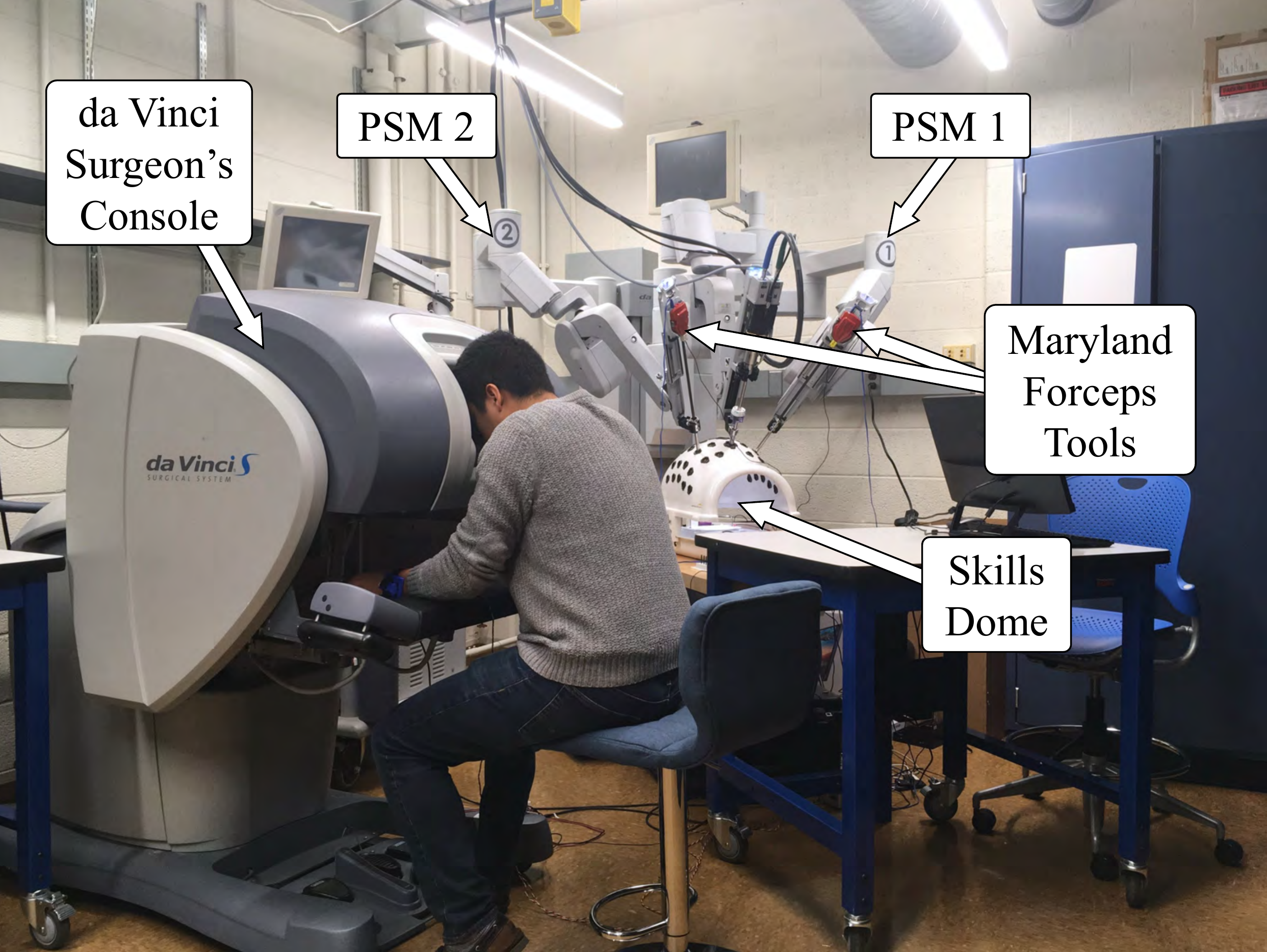}
    \caption{Participant seated at the da Vinci console to control the instruments interacting with the task materials inside the skills dome.}
    \label{fig:DAVINCI_OVERVIEW}
    \vspace{-10pt}
\end{figure}

Prior to each session, the experimenters set the camera view and initial positions of the da Vinci instruments. After providing informed consent, participants were randomized into either the feedback or no-feedback group. Participants in the feedback group wore the wrist-squeezing devices and received haptic feedback throughout the experiment. Participants in the no-feedback group did not wear the wrist-squeezing devices and received no haptic feedback. After being assigned to a group, participants completed a demographics survey and sat at the da Vinci console as shown in Fig. \ref{fig:DAVINCI_OVERVIEW}.

Participants were introduced to the da Vinci surgical system and received instructions on how to operate the controls. For participants in the feedback group, the wrist-squeezing tactile actuators were secured to participants' wrists and activated. The strap was then adjusted to ensure participants felt the entire range of squeezing sensations. Participants then had five minutes to practice grasping and manipulating objects using the da Vinci and a peg transfer training task \cite{Vassiliou2010FLSAssessment}. Participants were encouraged to explore the entire workspace with the surgical instruments. Following the warm-up session, the experimenters replaced the peg transfer board with the ring rollercoaster task, as shown in Fig.~\ref{fig:TASK_PLATFORM}, and explained the task instructions. 

In the ring rollercoaster task, the ring begins at the \emph{start} position on the left-hand side of the track, and participants were instructed to move it to the \emph{end} position on the right-hand side of the track. The \emph{start} and \emph{end} positions are each located halfway along the straight sections of the track. There are two curved sections and one straight section between the two positions. These positions were chosen such that the task could be completed without having to adjust the camera or use the clutch. Participants were required to pick the ring up with the left surgical instrument and perform at least one ring hand-off between left and right instruments during each trial. Participants were instructed to complete the task as quickly as possible, while centering the ring on the track. Participants were also told that the experimenters would notify them if excessive forces were being applied on the track. Participants wore headphones playing pink noise to mask any potential auditory cues from the haptic feedback devices or robot actuators. Each participant performed the ring rollercoaster task twelve times, for a total of 240 trials amongst all participants.

\subsubsection{Survey}

The twelve trials of the ring rollercoaster task were broken into four blocks of three trials each. Participants took a five-minute break after each trial block and completed a survey based on the NASA Task Load Index (TLX) questionnaire \cite{Hart-Staveland1988}. The survey consisted of rating-scale (1--10) and open-ended questions related to task difficulty and perceived task performance, as shown in Table \ref{SurveyQuestions}. The rating-scale survey questions were analyzed using statistical modeling. During this break, the experimenter also reminded participants of the task objectives. 

\begin{table}[!t]
\renewcommand*\arraystretch{1.1}
\caption{Survey questions}
\label{SurveyQuestions}
\begin{tabular}{p{0.03\linewidth} p{0.7\linewidth} p{0.12\linewidth}}
     \hline
     $\#$ & Question & Answer options\\
     \hline
     1.  & How mentally demanding was the task? & 1-10\\
     2.  & How physically demanding was the task? & 1-10\\
     3.  & Was there a change in your strategy / how you chose to manipulate the ring? Please describe. & -\\
     4.  & How successful were you at accomplishing your goal? & 1-10\\
     5.  & What prevented you from accomplishing your goal, if anything? & -\\
     6.  & How natural was your manipulation of the tools? & 1-10\\
     7.  & How frustrated, stressed, or annoyed were you? & 1-10\\
     8.  & How well could you concentrate on the task? & 1-10\\
     9.  & Other notes you want us to know? & -\\
     \hline
\end{tabular}
\vspace{-10pt}
\end{table}

\subsection{Metrics and Data Analysis}

To evaluate performance, two metrics were considered for each trial: task completion time and root-mean-square (RMS) force. As a measure of task speed, task completion time was calculated as the time between the first and last instances of contact with the ring rollercoaster track. For each trial, these instances were manually verified by watching the video recording of the trial. As a measure of task accuracy, RMS force was calculated as the RMS of the force magnitude signal calculated from the time-series data collected by the force/torque sensor for each trial. All recorded data were transformed using the base-10 logarithm, and the log task completion time and log RMS force were each visually determined to be normally distributed for each trial.

We used random-coefficients linear mixed-effects (LME) models to compare the log task completion time and log RMS force between feedback and no-feedback groups. Separate LME models were created for each metric; within each model, feedback condition and trial number were modeled as fixed effects and participant was modeled as a random effect. We also used random-intercepts LME models to compare the rating-scale responses from the NASA-TLX questionnaire between feedback and no-feedback groups. General linear hypothesis testing (GLHT) was used to assess the differences in rating-scale responses at the start of the experiment (trial 1). All statistical analyses were completed using R 3.6.3 and MATLAB R2020a; we use $\alpha=0.05$ to determine significance. \looseness=-1

\begin{figure}[!t]
    \centering
    \includegraphics[width=\columnwidth]{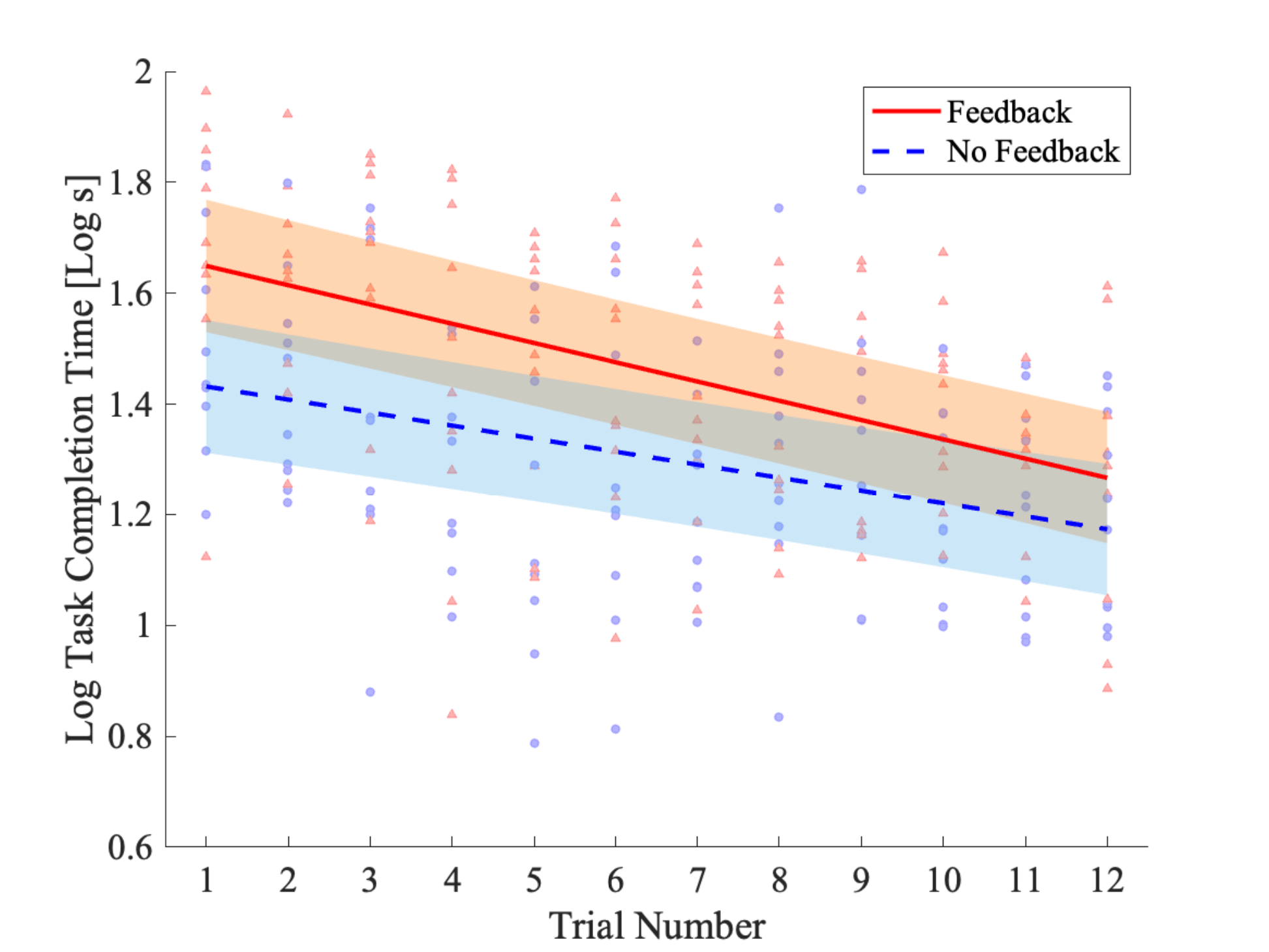}
    \caption{\small Log-transformed task completion time by feedback condition. Red triangular markers and solid fit line represent the feedback group. Blue circular markers and dashed fit line represent the no-feedback group.}
    \label{fig:LME_DURATION}
    \vspace{-10pt}
\end{figure}

\begin{figure}[!t]
    \centering
    \includegraphics[width=\columnwidth]{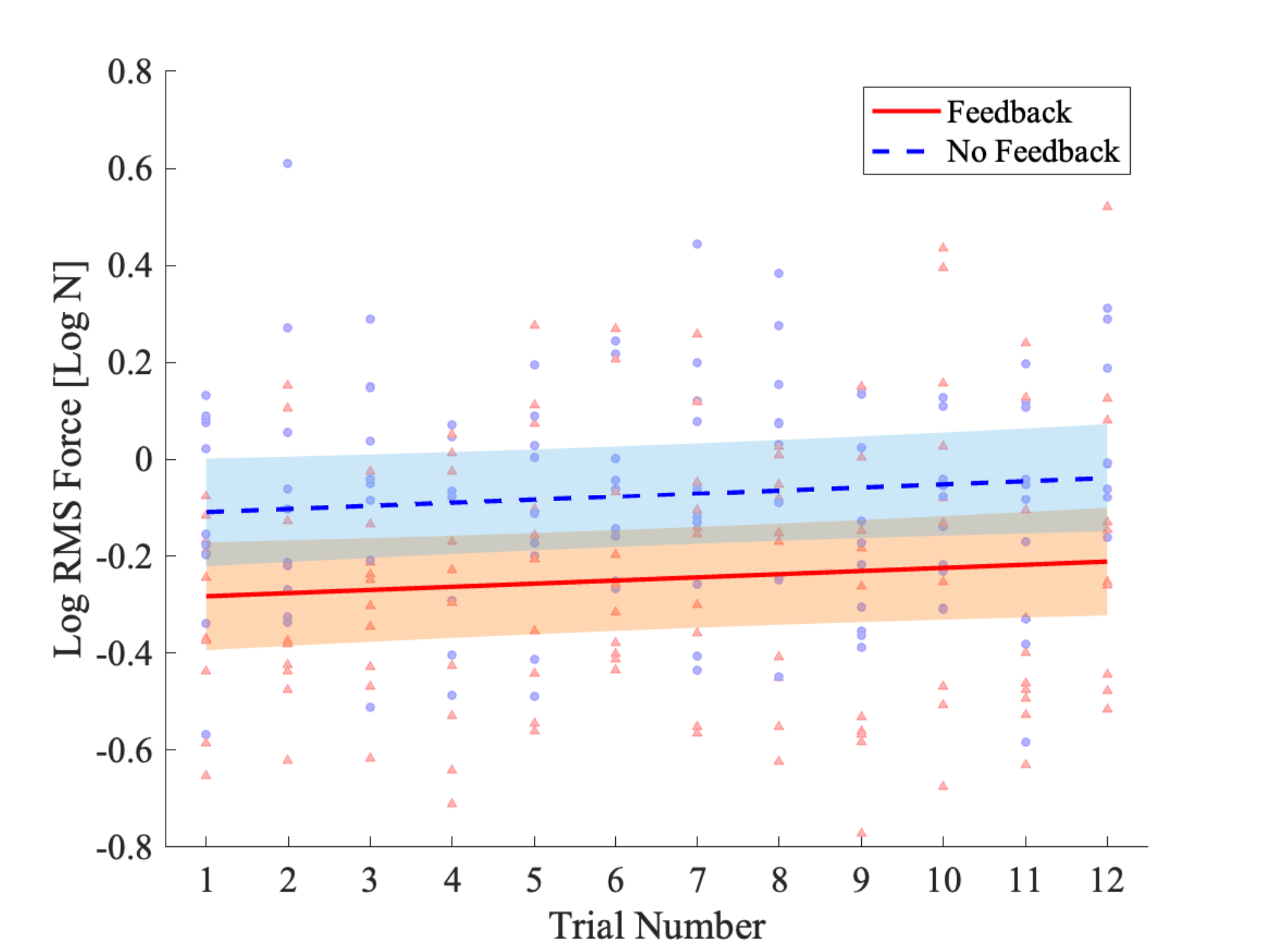}
    \caption{\small Log-transformed RMS force by feedback condition. Red triangular markers and solid fit line represent the feedback group. Blue circular markers and dashed fit line represent the no-feedback group.}
    \label{fig:LME_RMSFORCE}
    \vspace{-10pt}
\end{figure}

\section{Results}
Of the 240 trials, one was removed due to experimenter error (not starting data collection), and five were removed due to a system error in zeroing the force sensor. The following results are for the remaining 234 trials.

\subsection{Task Completion Time}

\begin{table}[!t]
\begin{center}
\renewcommand*\arraystretch{1.1}
\caption{Fixed effect results for the log task completion time LME model. NF is the intercept of the no-feedback group, FB is the intercept of the feedback group, TN is the trial number, and FB:TN is the interaction between feedback condition and trial number.}
\label{LogDurationLMETable}
\begin{tabular}{cccc}
     \hline
     Comparison & $\beta$ & SE & p-value  \\
     \hline
     NF    & $1.455$  & $0.062$ & $<2\mathrm{e}{-16}$ \\
     FB    & $0.229$  & $0.088$ & $0.014$ \\
     TN    & $-0.023$ & $0.004$ & $3.29\mathrm{e}{-10}$ \\
     FB:TN & $-0.011$ & $0.005$ & $0.026$ \\
     \hline
\end{tabular}
\end{center}
\vspace{-14pt}
\end{table}

Results of the LME model for task completion time showed significant fixed effects of feedback condition ($p=0.01$) and trial number ($p<0.01$), and a significant interaction effect between feedback and trial number ($p=0.03$). GLHT of the difference in log task completion time between feedback and no-feedback groups at Trial 1 demonstrated a significantly higher initial task completion time for the feedback group ($\beta=0.217$, $SE=0.086$, $p=0.01$). The no-feedback group significantly reduced their log task completion time over the twelve trials of the experiment ($p<0.01$). However, as shown in Fig. \ref{fig:LME_DURATION}, the feedback group reduced their log task completion time significantly faster than the no-feedback group ($p=0.03$). After back-transform\-ing the log task completion time data, we found that the feedback group decreased their task completion time by 7.68\% from trial to trial on average, while the no-feedback group decreased their task completion time by 5.26\% from trial to trial. These fixed effect results for log task completion time are summarized in Table \ref{LogDurationLMETable}.

\subsection{RMS Force}

\begin{table}[!t]
\begin{center}
\renewcommand*\arraystretch{1.1}
\caption{Fixed effect results for the log RMS force LME model. NF is the intercept of the no-feedback group, FB is the intercept of the feedback group, and TN is the trial number.}
\label{LogRMSForceLMETable}
\begin{tabular}{cccc}
     \hline
     Comparison & $\beta$ & SE & p-value  \\
     \hline
     NF    & $-0.116$ & $0.058$ & $0.055$ \\
     FB    & $-0.173$ & $0.075$ & $0.031$ \\
     TN    & $0.007$  & $0.004$ & $0.077$ \\
     \hline
\end{tabular}
\end{center}
\vspace{-10pt}
\end{table}

Results of the initial LME model for log RMS force showed a significant fixed effect of feedback condition ($p=0.02$). However, the fixed effect of trial number and the interaction effect of feedback and trial number were not significant ($p>0.05$), indicating that neither group significantly increased or decreased their log RMS force over the twelve trials. Given the lack of significant effects in this model, a simpler, random-intercepts model was analyzed that considered only the fixed effects of trial and feedback. The results of GLHT on this model showed that the feedback group had a significantly lower RMS force than the no-feedback group at Trial 1 ($\beta=-0.174$, $SE=0.075$, $p=0.02$), which was maintained over all twelve trials (see Fig.~\ref{fig:LME_RMSFORCE}). After back-transform\-ing the log RMS force data, we found that the feedback group was exerting approximately 0.67\,N less than the no-feedback group throughout the experiment. These fixed effect results for log RMS force are summarized in Table \ref{LogRMSForceLMETable}.

\subsection{Surveys}

\begin{table}[!t]
\begin{center}
\renewcommand*\arraystretch{1.1}
\caption{Mean ($\mu$) and standard deviation ($\sigma$) of survey rating responses for feedback (FB) and no-feedback (NF) groups.}
\label{Survey_rating_response_tbl}
\begin{tabular}{c p{4cm} c c c c }
    \hline
    \# & Question & \multicolumn{2}{c}{FB} & \multicolumn{2}{c}{NF} \\
    & & $\mu$ & $\sigma$ & $\mu$ & $\sigma$ \\  
    \hline
    1 & How mentally demanding was the task? & $3.6$ & $2.6$ & $2.6$ & $2.3$ \\
    2 & How physically demanding was the task? & $3.1$ & $2.2$ & $2.4$ & $1.9$ \\
    4 & How successful were you at accomplishing your goal? & $6.6$ & $2.2$ & $7.4$ & $2.2$ \\
    6 & How natural was your manipulation of the tools? & $6.2$ & $2.4$ & $7.2$ & $2.2$ \\
    7 & How frustrated, stressed, or annoyed were you? & $2.8$ & $2.3$ & $1.5$ & $0.7$ \\
    8 & How well could you concentrate on the task? & $8.4$ & $2.3$ & $9.3$ & $0.6$ \\ 
    \hline
\end{tabular}
\end{center}
\vspace{-16pt}
\end{table}

Table \ref{Survey_rating_response_tbl} shows the mean and standard deviation of the rating responses to each survey question for both groups.

Results of the LME models for Questions 1 and 2 showed a significant fixed effect of trial number ($p<0.01$) with a negative-valued estimate ($\beta$), indicating that the physical and mental demand decreased significantly for both groups over the course of the experiment. GLHT demonstrated no significant difference in physical or mental demand between the feedback and no-feedback groups at Trial 1 ($p>0.05$). Since the fixed effect of feedback condition was not significant ($p>0.05$), the feedback group did not report significantly different physical or mental demand compared to the no-feedback group throughout all trials.

Question 4 assessed how successful the participants were at accomplishing their goal. The results of the corresponding LME model showed a significant fixed effect of trial number with a positive-valued estimate, which indicates that the level of success experienced by the participants increased significantly for both groups throughout the experiment. Using GLHT, there was no significant difference in the success ratings between the two groups at Trial 1 ($p>0.05$). This result was held across all trials since the LME model results showed that the fixed effect of feedback condition was not significant ($p>0.05$).

The results of the LME model for Question 6 demonstrated a significant fixed effect of trial number with a positive-valued estimate, indicating that the participants' rating of ``How natural was your manipulation of the tools?'' increased significantly for both groups over the experiment. Again, there was no significant difference in perception of natural manipulation between feedback and no-feedback groups at Trial 1, using GLHT ($p>0.05$). Since the fixed effect of feedback condition was not significant ($p>0.05$), the result at Trial 1 did not change over the trials.

The LME model corresponding to Question 7, ``How frustrated, stressed, or annoyed were you?'', showed a significant fixed effect of trial number ($p<0.02$) with a negative-valued estimate, indicating that the frustration level decreased significantly for both groups over the course of the experiment. GLHT demonstrated no significant difference in frustration ratings between the feedback and no-feedback groups at Trial 1 ($p>0.05$). Again, the fixed effect of feedback condition was not significant ($p>0.05$), so the feedback group did not report significantly different frustration with the task than the no-feedback group over all trials. 

For Question 8 (``How well could you concentrate on the task?''), responses were clustered around the higher end of the rating scale (1-10), as indicated by the means and standard deviations shown in Table \ref{Survey_rating_response_tbl}. The results of the LME model for concentration ratings demonstrated that the ratings did not increase or decrease significantly for either group throughout the experiment ($p>0.05$). Similar to the results of the other LME models for survey responses, there was no significant difference in ratings between the feedback and no-feedback groups at Trial 1 (using GLHT), which held throughout all trials since the fixed effect of feedback condition was not significant ($p>0.05$).

\section{Discussion}
In this study, we sought to investigate how providing people with haptic feedback of the forces they apply during a RMIS training task would impact their performance, particularly regarding the speed-accuracy tradeoff. We hypothesized that providing novice trainees with tactile feedback of the forces they are applying would significantly reduce their interaction force during these training activities while also increasing their task completion time, compared to conventional training without haptic feedback. The presented results indicate that participants receiving haptic feedback applied significantly less force throughout the experiment while achieving final task completion times that are not different from participants receiving no haptic feedback.

\subsection{Accuracy Results}
Our finding that haptic feedback improves force-based task performance is consistent with results from prior studies investigating haptic feedback \cite{Brown2017,Abiri2019ArtificialFeedback,Abiri2019Multi-ModalSurgery,Wottawa2013TheTasks,King2009TactileSurgery}. For example, using a simpler version of our wrist-squeezing feedback system that did not have bimanual capabilities, Brown et al.\ showed that receiving force feedback helped participants significantly reduce the integral of force they produced throughout the experiment \cite{Brown2017}. While their study highlighted the potential residual effect of haptic feedback, the performance improvements were never compared to a control group. Our results complement this prior work by demonstrating that haptic feedback helped the experimental group perform the training task with lower forces than the control group. In a manner similar to our findings, King et al.\ achieved a decrease in grip force compared to the no-feedback group \cite{King2009TactileSurgery} in a peg transfer task. However, they did not report the impact this improvement had on the task completion time. It is also worth noting that their haptic feedback approach required significant modifications to the surgical instruments and robot controls. All of the sensing and actuation components used in our approach are external to the robotic platform, which allows for generalization to other training tasks. 

\subsection{Task Completion Time Results}
Complementary to our finding of differences in force production between our two groups, we also highlighted the impact of haptic feedback on the speed at which participants completed the task. Although the tradeoff between speed and accuracy in motor tasks is well known \cite{Meyer1982ModelsMovements}, to the best of our knowledge, no single study has examined the effect of haptic feedback on this tradeoff for RMIS training. Several studies have looked at task completion time as a task performance metric \cite{Hubens2003ASystem,Bark2012SurgicalTasks,Brown2017}. For example, Chien et al.\ found a significant linear correlation between movement time and index of difficulty in a grasp-and-hold experiment \cite{Chien2010AccuracySurgery}. Speed and accuracy are both widely used to assess performance in RMIS training tasks, typically individually. Our results are unique since they demonstrate that haptic feedback has a significant impact on the complex interaction between speed (task completion time) and accuracy (RMS force) in RMIS training. Though the participants in the feedback group were significantly slower than the participants in the no-feedback group at the beginning of the study, by the end of the twelve trials, the task completion time was not significantly different between the two groups. Not only did haptic feedback reduce the task completion time at a significantly faster rate for the feedback group, but it also allowed the feedback group to maintain significantly lower interaction forces from the beginning to the end of the study. While we cannot describe how performance on this tradeoff changes after twelve trials, we can definitively say that any initial negative impact of haptic feedback on task completion time was attenuated with repeated practice, in much the same way that speed in the control group improves with practice. 

\subsection{Survey Results}
Our qualitative survey results indicate that there was no significant difference in physical or mental demand, how successful the participants were at accomplishing their goal, how natural the manipulation of the tools was, how frustrated, stressed, or annoyed the participants were, and how well the participants could concentrate on the task between participants receiving haptic feedback and those that did not. Furthermore, the success ratings, natural manipulation ratings, and concentration ratings increased significantly over the study, while the perceived physical demand, mental demand, and frustration ratings decreased significantly throughout the experiment for all participants. These results suggest that supplying novice surgeons with continuous haptic feedback of the forces they apply during training does not present any hindrance compared to completing the task without haptic feedback; indeed our quantitative results show that the studied form of haptic feedback can even help to improve their ability to perform the task. As expected for feedback that points out mistakes, a small number of participants mentioned that the feedback was annoying because it captured their attention and caused them to focus on accuracy over speed. Participants in the haptic feedback group also did not report any other adverse effects of wearing the devices or receiving wrist-squeezing feedback while performing the task. \looseness=-1

\subsection{Limitations}
While our experimental findings further substantiate the benefits of RMIS training with haptic feedback, there are a few limitations that should be noted. First, our experiment utilized novice participants from the general population. We do not expect this to have a significant impact on our results because the training task can be performed accurately with no clinical knowledge. At the same time, it would be ideal to validate and expand on these findings with a participant pool of surgical trainees. Second, our results hold only for the ring rollercoaster task. Many previous studies \cite{Abiri2019Multi-ModalSurgery,King2009TactileSurgery,Wottawa2013TheTasks} used a peg transfer task, which is a validated component of the Fundamentals of Laparoscopy (FLS) curriculum \cite{Vassiliou2010FLSAssessment}. Even though both tasks are commonly used in RMIS training, they lack the clinical relevance of other surgical tasks like dissection and suturing.

Next, while we expect the handedness determination circuit would function during clinical procedures, we have not tested it in such a setting. In contrast, the force sensor beneath the task materials is not compatible with surgery on human patients. This approach to haptic feedback in RMIS is thus suitable only for training with conductive inanimate materials or {\em ex vivo} tissue. Also, the restriction we placed on camera movement differs from clinical practice, where surgeons frequently change the camera position by clutching (by means of a foot pedal or finger clutch). Finally, our experiment ended after twelve trials and is therefore unable to assess how the observed trends change longitudinally. Although we hypothesize that the two groups would plateau at different levels \cite{Maniar2005ComparisonAdoption}, this hypothesis needs to be formally tested. \looseness=-1

\subsection{Future Work}
Future investigations will evaluate the utility of haptic feedback over longer time periods and in clinically relevant tasks such as cutting and suturing \cite{Koehn2015a,Bark2012SurgicalTasks}. To this end, we will recruit surgical trainees, record their performance, and provide one or more modalities of haptic feedback (including the tested modality of wrist-squeezing force feedback) using our new data acquisition/haptic feedback framework. Also, to one day enable the utility of our system in a real surgical setting, we will consider using external force estimation methods such as the one developed by Yilmaz et al. \cite{Yilmaz2020NeuralKit}. Finally, we will upgrade the wrist-squeezing device to use a brushed DC motor instead of a servo; this will allow us to test higher fidelity force feedback methods such as those developed by Pezent et al. \cite{Pezent2019Tasbi:Reality}.

\section{Conclusion}
In this study we augmented a da Vinci S surgical system with bimanual wrist-squeezing haptic feedback to investigate the effect of haptic feedback on early RMIS training for novice participants. The system provides real-time haptic feedback of the forces applied on a surgical training task, as measured by a force/torque sensor located underneath the task board. We found that the haptic feedback of interaction forces significantly changes the speed and accuracy with which novices complete a ring rollercoaster training task. In particular, haptic feedback caused novices to complete the task less quickly during their first trials, but more accurately throughout all trials. Over repeated trials, novices using haptic feedback maintained a superior level of accuracy and reached the speed achieved by novices in the control group. These results support half of our hypothesis: providing novice trainees with tactile feedback of the forces they apply while training significantly reduced their interaction force during training, compared to conventional training. However, regarding the other half of our hypothesis -- that trainees receiving haptic feedback would increase their task completion time -- we instead found that trainees significantly reduce their task completion time over the course of the study. This unexpected outcome in task completion time, along with the survey results, demonstrates that providing novice trainees with wrist-squeezing haptic feedback of their interaction forces creates no adverse effect on speed compared to training without haptic feedback. Instead, haptic feedback over twelve trials can help novice trainees reduce their interaction forces while finishing the task at a speed indistinguishable from trainees receiving no feedback. Overall, these results suggest that this type of haptic feedback could play an important role in the development of technical skills by novice robotic surgeons.

\section*{Acknowledgments}
We would like to thank Dr. Leah R. Jager for expert consultation on the statistical analyses in this study.

 


\bibliography{references.bib}
\bibliographystyle{IEEEtran}


\end{document}